\documentclass[preprint,journal]{vgtc}            % preprint (journal style)

\onlineid{0}

\vgtccategory{Research}

\vgtcpapertype{please specify}

\title{Cluster-Based Random Forest Visualization and Interpretation}

\author{%
  \authororcid{Max Sondag}{0000-0003-3309-638X},
  \authororcid{Christofer Meinecke}{0000-0002-5637-9975},
  \authororcid{Dennis Collaris}{0000-0001-7612-9319},
  \authororcid{Tatiana von Landesberger}{0000-0002-5279-1444}, and
  \authororcid{Stef van den Elzen}{0000-0003-1245-0503}
}

\authorfooter{
  \item
  	Max Sondag is with University of Cologne, Germany and Maastricht University, Netherlands.
  	E-mail: max.sondag@maastricht-university.nl
    \item
    Tatiana von Landesberger is with University of Cologne, Germany.
  	E-mail: tlandesb@uni-koeln.de
  \item
  	Christofer Meinecke is with Leipzig University and ScaDS.AI Dresden/Leipzig, Germany.
  	E-mail: cmeinecke@informatik.uni-leipzig.de

  \item Dennis Collaris and Stef van den Elzen are with Eindhoven University of Technology, the Netherlands.
  	E-mail: \{d.a.c.collaris, s.j.v.d.elzen\}@tue.nl.
}

\abstract{
 Random forests are a machine learning method used to automatically classify datasets and consist of a multitude of decision trees. While these random forests often have higher performance and generalize better than a single decision tree, they are also harder to interpret. This paper presents a visualization method and system to increase interpretability of random forests. We cluster similar trees which enables users to interpret how the model performs in general without needing to analyze each individual decision tree in detail, or interpret an oversimplified summary of the full forest. To meaningfully cluster the decision trees, we introduce a new distance metric that takes into account both the decision rules as well as the predictions of a pair of decision trees. We also propose two new visualization methods that visualize both clustered and individual decision trees: (1) The Feature Plot, which visualizes the topological position of features in the decision trees, and (2) the Rule Plot, which visualizes the decision rules of the decision trees. We demonstrate the efficacy of our approach through a case study on the ``Glass'' dataset, which is a relatively complex standard machine learning dataset, as well as a small user study.
}


\teaser{
  \centering
  \begin{overpic}[width=\linewidth]{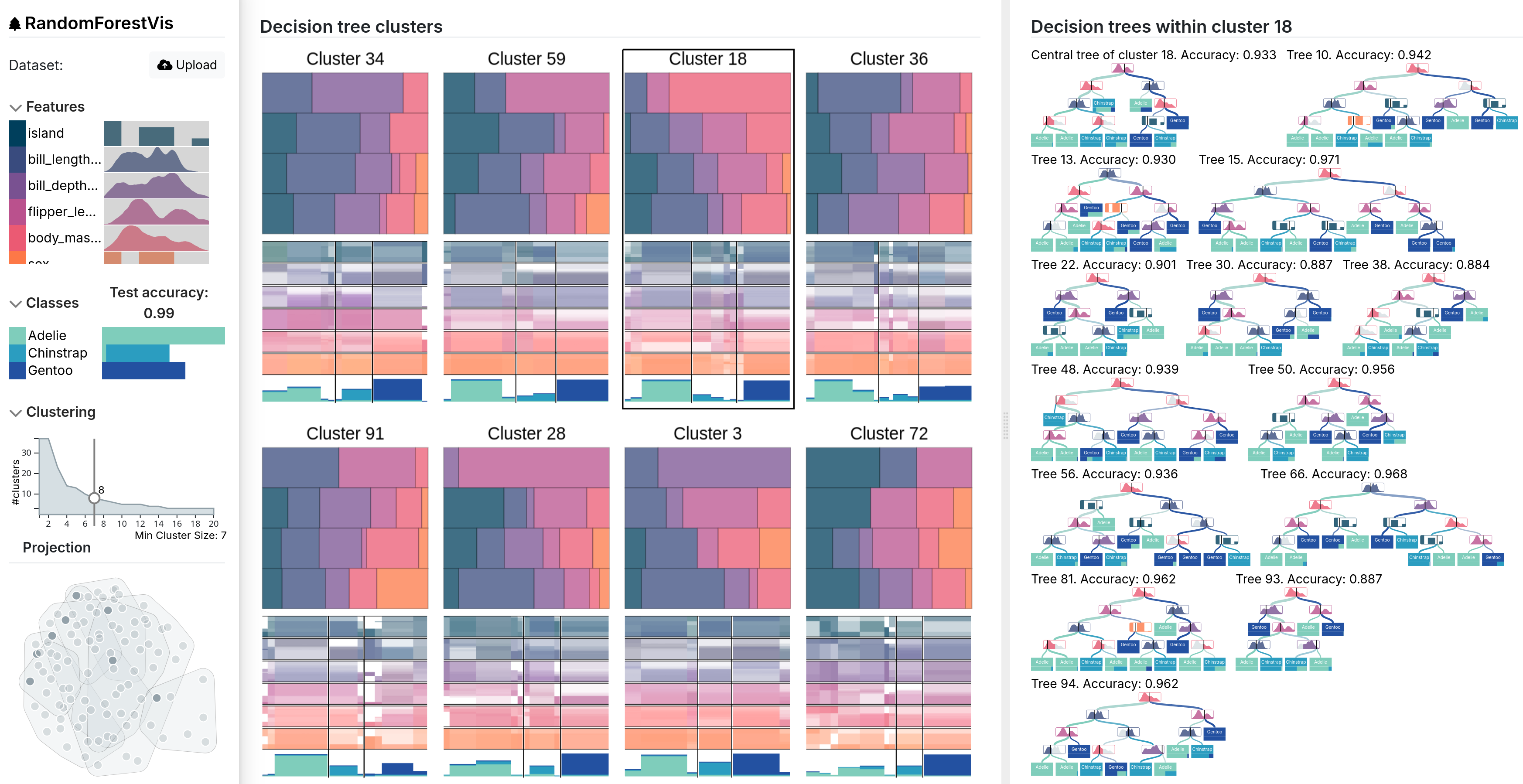}
    \put(11,48){
        \begin{tikzpicture}
        \draw[fill=black] (0,0) circle [radius=0.3] node[text=white] {A};
        \end{tikzpicture}
    } 
    \put(39,48){
        \begin{tikzpicture}
        \draw[fill=black] (0,0) circle [radius=0.3] node[text=white] {B};
        \end{tikzpicture}
    } 
    \put(95,48){
        \begin{tikzpicture}
        \draw[fill=black] (0,0) circle [radius=0.3] node[text=white] {C};
        \end{tikzpicture}
    } 
  \end{overpic}
  \caption{%
        An overview of the visual analytics system. A) shows global information such as feature and class distribution, model accuracy, and predictions. Furthermore, it allows control of the number of clusters that are displayed and shows a 2D projection of the clustered trees. B) The Feature Plot (top) shows how often a feature is used within a cluster at a specific depth in the tree and the Rule Plot (bottom) shows an aggregated view of the classification rules used in a cluster. C) Shows the decision trees within a selected cluster.
  }
  \label{fig:teaser}
}

\graphicspath{{figs/}{figures/}{pictures/}{images/}{./}} % where to search for the images

\usepackage{mathptmx}                  % use matching math font

\newcommand\interaction[1]{\textcolor{black}{#1}}
\usepackage{wrapfig}

\usepackage[T1]{fontenc}
\usepackage{overpic}
\usepackage{amsmath}
\newcommand\revision[1]{\textcolor[rgb]{0,0.0,0}{#1}}
\newcommand\buildingFloatColor[1]{\textcolor[rgb]{0.50,0.80,0.65}{#1}}
\newcommand\buildingColor[1]{\textcolor[rgb]{0.30,0.721,0.756}{#1}}
\newcommand\headlampsColor[1]{\textcolor[rgb]{0.11,0.19,0.53}{#1}}

\newcommand\billlength[1]{\textcolor[rgb]{0.38,0.44,0.59}{#1}}
\newcommand\billdepth[1]{\textcolor[rgb]{0.47,0.31,0.57}{#1}}
\newcommand\gentooColor[1]{\textcolor[rgb]{0.13,0.31,0.63}{#1}}
\newcommand\adelieColor[1]{\textcolor[rgb]{0.48,0.79,0.73}{#1}}

\newcommand\refractiveindexcolor[1]{\textcolor[rgb]{0.26,0.43,0.52}{#1}}
\newcommand\sodiumcolor[1]{\textcolor[rgb]{0.14,0.28,0.45}{#1}}
\newcommand\magnesiumcolor[1]{\textcolor[rgb]{0.31,0.31,0.53}{#1}}
\newcommand\calciumColor[1]{\textcolor[rgb]{0.92,0.35,0.45}{#1}}
\newcommand\bariumcolor[1]{\textcolor[rgb]{0.98,0.42,0.34}{#1}}
\newcommand\ironColor[1]{\textcolor[rgb]{0.98824,0.63922,0.39216}{#1}}

\begin{document}
%%%%%%%%%%%%%%%%%%%%%%%%%%%%%%%%%%%%%%%%%%%%%%%%%%%%%%%%%%%%%%%%
%%%%%%%%%%%%%%%%%%%%%% START OF THE PAPER %%%%%%%%%%%%%%%%%%%%%%
\maketitle

\section{Introduction}
Machine learning models are used in many application areas, from personalized recommendation systems to large-scale decision systems in industry~\cite{sarker2021machine}.
This presence of machine learning comes with challenges regarding the black-box nature of these models, including issues of transparency, interpretability, and explainability~\cite{chatzimparmpas2020state}.

In the case of tabular data, Random Forest (RF)~\cite{breiman2001random} and other tree-based models are popular and effective classification methods that still outperform neural networks~\cite{grinsztajn2022tree}.
RF is an ensemble learner composed of several decision trees, but despite their utility, the complexity of ensemble models can obscure the understanding of how they arrive at their predictions.
It may be possible to interpret a single decision tree~\cite{Barlow2001EMTree, Soon2003PaintingClass, Elzen2011BaobabView}, or to compare a small number of decision trees quite easily, but with increasing number of features and trees, this becomes challenging~\cite{vidal2020born, sagi2020explainable}.
Existing work on RF explainability mainly focuses on feature contributions (e.g., ~\cite{zhao2018iforest, Macas2024RF}), extracts decision rules (e.g., ~\cite{chatzimparmpas2023visruler, neto2020explainable}), or aims to create a single tree summary from the whole ensemble (e.g., ~\cite{vidal2020born, sagi2020explainable}). Typically, all rules are combined into a single representation of the full classification model, or the individual trees are presented in detail. Both methods are challenging to interpret and understand, and hide the multi-level structures present in the model. We believe an RF model is not best explained with a single summary, but through clusters of trees that are responsible for classification of data with specific characteristics. Understanding the different clusters in the model enables users to see how well specific parts of the data are covered by different trees. This provides insight into the robustness and confidence of the model by inspecting the cluster sizes, feature characteristics, and both common and outlier trees.

\medskip
We propose an approach that aims to find a middle ground between visualizing the entire model and visualizing a single tree to explain the model. For this, we leverage the structure in the decision trees to cluster these based on their rules (paths).
An important aspect of this approach is the selection of an appropriate distance metric. 
We focus on the rules, which allow us to include \emph{semantic} as well as \emph{structural} components of the trees.
The final cluster-based visualization of the model should reduce the cognitive load while remaining (largely) faithful to the original model. 

\medskip
For the cluster-based visualization and interpretation of an RF model, we propose a visual analytics system that uses three levels of granularity.
First, we show global information about the model and the dataset, showing the overall classification results, accuracy, and feature distribution (see Figure~\ref{fig:teaser}a).
Secondly, we create two visualizations that aggregate the features and the rules of the decision trees in a cluster.
By distilling the complexity of RFs into a manageable set of trees, this approach can help to understand these models (see Figure~\ref{fig:teaser}b).
Lastly, we include a detailed view that shows all decision trees in a cluster, allowing users to inspect the behavior of individual decision trees on their dataset (see Figure~\ref{fig:teaser}c).
Insights using our system help a variety of applications, ranging from diagnostics, refinement, decision support, and justification.
For evaluation purposes, we present a case study on the ``Glass'' dataset\cite{glass_identification_42} and conduct a small user study in a think-aloud setting.
For this, we collect responses with the System Usability Scale (SUS)~\cite{brooke1996sus} and ICE-T~\cite{wall2018heuristic} questionnaires.

Our contributions are as follows:

\vspace{1em} 
\begin{enumerate}
\item A new distance metric for comparing decision trees in a random forest based on both the predictions and decision rules \emph{(semantics and structure)}.
\item Two new scalable aggregation-compatible visualizations for exploring random forests based on their features and decision rules.
\item A cluster-based visualization system for visualizing a random forest model using aggregation.
\end{enumerate}
\vspace{1em}

The paper is structured as follows: in Section~\ref{sec:related_work} we discuss work related to the visualization of decision trees, random forests, and distance metrics for decision trees. Section~\ref{sec:tasks} elicits the tasks we aim to support with our approach. Next, we introduce and discuss our distance metric and the associated method of clustering the decision trees in Section~\ref{sec:distancemetric}. We then use these methods in our interactive visualization approach presented in Section~\ref{sec:vis}. Section~\ref{sec:evaluation} evaluates our approach with both case and user studies. Finally, limitations, directions for future work, and conclusions are presented in Section~\ref{sec:discussion} and Section~\ref{sec:conclusion}, respectively.

\section{Related work}
\label{sec:related_work}
Central to our method for the visualization and interpretation of random forests are the visualization of clusters of decision trees that together form the random forest, and the visualization of individual decision trees. To construct the clusters, we group together similar decision trees. For this, we need a distance metric to determine the similarity between the trees. In this section, we discuss work related to these elements; visualization of decisions trees; visualization of random forests; and proposed distance metrics for decision trees.

\subsection{Visualization of Decision Trees}
As decision trees represent a hierarchy, standard hierarchical visualization techniques and guidelines are applicable, as proposed by Elmqvist and Fekete~\cite{Elmqvist:2010:HierarchicalAggregation}. The tree structure can be encoded explicitly or implicitly as surveyed by Schulz et al.~\cite{Schulz:2010:Survey}. Graham and Kennedy~\cite{graham2010survey} provide a survey on the visualization of multiple trees simultaneously, further explored for multivariate data by Zheng and Sadlo~\cite{Zheng2021hierarmultivis}. Building on this, several visualization and interaction techniques have been developed to compare tree representations~\cite{Munzner2003TreeComparison, Bremm2011TreeComparison, Liu202TreeComparison}). However, these standard hierarchical data visualizations, as well as model-agnostic machine learning visualizations (e.g., Confusionflow~\cite{hinterreiter2020confusionflow}, and ModelWise~\cite{Meng2022ModelWise}), do not leverage the unique properties of decision trees such as features used, split rules, and classification data flow. 

\medskip
Split rules are visualized using different idioms such as icicle plots~\cite{ankerst2000icicle, Liu2007icile}, parallel coordinate plots~\cite{han2000ruleviz}, and adapted treemaps~\cite{chen2008visualizing}. Li et al.~\cite{li2019barcodetree} introduce a compact, barcode-inspired layout that enables the comparison of multiple tree structures by encoding the tree with a single barcode. However, for these paradigms it is more difficult to understand the tree structure (e.g., paths), as this is not directly visualized.

\medskip
EMTree~\cite{Barlow2001EMTree} uses a multiple linked view approach using treemaps, line-charts, bar-charts, and a node-link diagram to visualize the tree structure. PaintingClass~\cite{Soon2003PaintingClass} shows the tree structure with a focus-and-context approach, always having a single node in focus represented by star- or parallel-coordinate plots. Ancestors of the node are drawn smaller to the right, and descendants at the bottom. \revision{BaobabView}~\cite{Elzen2011BaobabView} integrates the tree visualization with the data visualization by explicitly showing the data as size- and color-encoded Sankey flows on top of the tree structure. Also, Worland et al.~\cite{worland2022visualization} visualize the tree structure by visualizing the classification boundaries at each level using a scatterplot and connecting linked data items. However, this method is not space efficient and introduces clutter due to many crossing edges, making it challenging to interpret the tree structure and understand the splits.

\medskip
For a more extensive and recent discussion of tasks and visualizations tailored for decision trees and rule-based classifiers, we refer the reader to the overview presented by Streeb et al.~\cite{streeb2021task}.

\subsection{Visualization of Random Forest}
\label{sec:relatedWorkRandomForest}
As random forests consist of (potentially a large number of) decision trees, visualizing them as a whole requires aggregation techniques. However, visualization of individual trees is possible if few trees are involved, and the trees themselves are rather simple with a low depth. Trees with a low depth can in general be achieved using different pruning techniques~\cite{kulkarni2012pruning}. In Médoc et al.~\cite{medoc2022visualizing} shallow trees (of maximum depth 5) are visualized using icicle plots positioned in a grid. Further linked visualizations provide additional interpretation mechanisms, such as a bar chart of feature importances and a scatterplot showing the residual versus prediction for all regression trees. Also, TreePOD~\cite{muhlbacher2017treepod} visualizes individual decision trees using pixel-based treemaps. The treemaps show the (implicit) tree structure, where each class-colored pixel in the contained rectangles represents a single data item. The system enables users to explore the relation between the number of nodes and the accuracy through a Pareto frontier of trees, guiding them to select the Pareto optimal tree. Individual trees can be inspected in detail using the flow-based tree visualization of BaobabView~\cite{Elzen2011BaobabView}. 

\medskip
% Rules based
As both the number of trees and their depth are typically large, most visualization techniques rely on summarization and aggregation of different aspects of the decision trees~\cite{aria2021comparison}. Wang et al.~\cite{wang2022timbertrek} present TimberTrek that visualizes a Rashomon set of more than a thousand decision trees using a radial icicle plot. In the radial plot, the rules of the shallow decision trees are combined into a single hierarchical representation. The individual trees can be explored in detail using additional views with node-link diagrams. Next to visualizing extra information and aggregation (size reduction), the other main interpretability strategies for random forests are rule extraction and local feature-based explanations~\cite{aria2021comparison}.

\medskip
Rulematrix~\cite{ming2018rulematrix} visualizes rules derived from the decision trees as different flows, horizontally branching from a main vertical data flow. Antweiler and Fuchs visualize rules together with a hierarchical radial node-link diagram representing the trees~\cite{antweiler2022visualizing}. Each rule is visualized separately using a block diagram, potentially leading to scalability concerns as the number of rules can be large for a moderate-size random forest, as decision trees typically need to cover for different characteristics of the data. VisRuler~\cite{chatzimparmpas2023visruler} addresses this by presenting all rules in a single view using an adapted parallel coordinates plot. They present a visual analytics system for extracting decision rules and use dimensionality reduction to project each decision tree classifier to a point. Another solution to deal with scalability is by reducing the ruleset to ``common'' rules. Adilova et al.~\cite{adilova2023re} focus on finding the best trimmed ruleset and present ways to visualize these. Neto and Paulovich~\cite{neto2020explainable} present an Explainable Matrix approach where rows are rules, columns are features, and cells are rules predicates, aiming to \revision{address} both scalability and interpretability. However, scalability remains a concern, as rules can be plenty, quickly rendering the rows too small.

\medskip
% Feature based
Next to rules-based visualizations, several methods focus on visualizing the feature contributions, e.g., using three-dimensional surfaces~\cite{welling2016forest} or using Sankey-like visualizations showing the flow of items through the different features, ordered by importance in the iforest system~\cite{zhao2018iforest}. This typically reveals the relationships between features and predictions and enables easier case-based reasoning.

\medskip
Finally, several methods focus on \emph{explainability} of the random forest. Aria et al.~\cite{aria2023explainable} summarize a random forest with a single decision tree. \revision{Such an approach provides valuable global insights that are not easily achieved through local analysis and visualization techniques. However, our approach has a different focus, aiming to identify and discover different clusters of decision trees in the random forest, striking a balance between a pure global and local approach.}
\revision{Other methods retrieve insights} on the \emph{data} rather than the \emph{model}, e.g., RfX~\cite{eirich2022rfx} demonstrates a visual analytics system for Random Forest interpretation. Mazumdar et al.~\cite{mazumdar2021random} present Random Forest Similarity Maps, addressing both scalability and interpretability of the model's inner workings by mapping the decision trees to 2D using dimensionality reduction techniques. In our Sidebar (Figure~\ref{fig:teaser}a), we use a similar projection for a quick overview of clusters of similar trees.

\medskip
As a general observation, most visualizations are relying on color for either the features and/or the classes involved, which limits the scalability of these. The dominant approach to visualize individual decision trees is through tree structured Sankey-like data flows~\cite{Elzen2011BaobabView}. For Random Forests, two methods are prevalent: rules extraction and feature based visualizations. For both, either the individual trees, and rules are shown, or an aggregate of all trees. Here we aim for a different approach, inspired by the work of Sondag et al.~\cite{Sondag2022Representative} we identify clusters of trees based on behavioral characteristics and visualize them using novel interactive summaries.

\subsection{Distance Metrics for Decision Trees}
Over the years, a multitude of metrics have been proposed to compute distances between decision trees.
Chipman et al.~\cite{chipman1998extracting} and Banerjee et al.\cite{banerjee2012identifying} proposed three metrics, 
one based on prediction agreement on a test dataset (actual class), one based on agreement in the leaf nodes on a test dataset (actual node), and one based on split points. 
However, both papers define the metric for split points differently. 

\medskip
%semantic, leaf agreement
Semantic methods only look at the classification results of the given data samples, e.g. two trees are similar if they predict the same classes for the same data samples or end up in the same leaf node. 
This computes an agreement score, but ignores the thresholds used at the split points.
Tan et al.\cite{tan2020tree} use a distance based on leaf agreement to select samples from the dataset that explain the prediction of a class.
Another way to compute distances based on a dataset is to compute feature contributions~\cite{palczewska2014interpreting}.
Thresholds are also ignored by methods that define two trees as similar if they use the same split variables~\cite{banerjee2012identifying}, which only shows differences if at least one variable is not used in a tree; otherwise, all distances are 0. 
These metrics do not fully capture the underlying tree structure.

\medskip
% structure, topology
The split method in Chipman et al.~\cite{chipman1998extracting} is based on Shannon and Banks method~\cite{shannon1999combining} and counts the number of nodes where the variable used for a split point is different with an optional penalty for the depth where the split occurs.
This is similar to the idea of Laabs et al.~\cite{laabs2023identification}, which defines two trees as similar if they use the same split variables, weighted by the depth and the number of times the variable is used. 
They showed that this approach can have advantages when choosing a representative tree over the methods of Banerjee et al.~\cite{banerjee2012identifying}.
A drawback of these methods is that they assign a non-zero distance to identical trees that differ only in the order of their split points, even when the thresholds remain the same. 
Another structural approach is to transform the node order into a string sequence, where two trees are similar if the same sequences of split variables are used to predict a class~\cite{eirich2022rfx,bakirli2017dtreesim}, which can be measured using the Levenshtein distance~\cite{levenshtein1966binary}.

\medskip
%rules, thresholds
None of these methods includes the thresholds of the split points.
In contrast, our approach is based on rules and includes the actual split values. Each rule represents a path from the root to a leaf and consists of multiple split points, each defined as an interval. For each rule, a set of intervals is computed on the basis of the split points, with an interval for each feature. 
Including the split values allows our method to capture the actual decisions made during the splitting process, rather than just the selected features, as in other methods. 
A similar approach to compare rules is used by Adilova et al.~\cite{adilova2023re} but, in contrast to them, we focus on tree distances and include interval overlap to ensure that cases that do not overlap are further apart. Other rule-based approaches focus on extracting logic rules from each path \cite{neto2020explainable, mazumdar2021random}, but do not use intervals to calculate distances.

\section{Task elicitation}
\label{sec:tasks}

A variety of different visualizations for random forests and associated tasks have already been explored in Section~\ref{sec:relatedWorkRandomForest}. Many of these tasks are from the perspective of machine learning experts who develop models~\cite{zhao2018iforest,medoc2022visualizing}, from the end-user who wants to understand how a single instance decision was made~\cite{zhao2018iforest,aria2021comparison,antweiler2022visualizing}, or from users of the model who want to generate a simplified model out of it~\cite{eirich2022rfx,aria2023explainable,aria2021comparison,chatzimparmpas2023visruler,muhlbacher2017treepod,medoc2022visualizing,wang2022timbertrek}. In this section, we elicit tasks from practitioners who use decision forests in their applied research for data classification purposes, but who are not machine learning experts themselves.

\medskip
We use a semi-structured interview setup to determine how two experts from the domains of computational modeling and geography are using random forests in their applied research. 
In particular, we are interested in what analysis tasks they are using random forests for, what their current process and focus is, and what they would like to be able to do and achieve in an ideal case. The complete list of questions and anonymized transcripts are available in the supplementary material.

\medskip
In order to structure the results of the interview, we used the categorization by Streeb et al.~\cite{streeb2021task} for the tasks. The main category on which our interviewees focus when using \revision{random} forests is on the \textit{validation} and \textit{evaluation} of the model: how good is the classifier working in general? Are there classes or areas where it performs better or worse? Is it reliable enough to be applicable? Are known salient split points used?
\revision{We also} identified a desire to better \textit{understand} the model: how confident is the model in the results? What is the overall structure of the forest? Are there features that are particularly important for a class?  For the feature importance, they also express a desire to see the relations between features, as high correlations can indicate overrepresentation.

\medskip
Both the evaluation and the understanding serve as a measure of the trustworthiness of the model for the interviewees. They aim to establish that the model is trustworthy enough such that they can use it on real-world applications, and that there are no unexpected results or important classes/input clusters that are likely to be incorrect.

\medskip
Finally, both our interviewees are working with spatial data. Hence, they are quite interested in seeing the visualization results on the spatial output space as well. That is, seeing for regions in the space how reliable the model is. While important for these particular users, this is not a problem that generalizes over all the use cases for random forests, as many datasets do not have this spatial component. Hence, we do not address this directly in our system, but it may be interesting for spatial data in particular to provide a two-dimensional view that allows for the visualization and filtering of this data.

From the results of these interviews, as well as surveying related work, we aim to support the following tasks in our visualization. Users should be able to:
\begin{itemize}
\item[\textbf{T1}] understand the overall decision process of the random forest.
\item[\textbf{T2}] understand how a particular decision is made.
\item[\textbf{T3}] evaluate how well the model performs globally.
\item[\textbf{T4}] evaluate model performance on specific input classes/clusters.
\item[\textbf{T5}] analyze the importance of the features. 
\end{itemize}

We additionally require that the following general constraint is satisfied: the visualization should be a faithful representation of the underlying model to generate trustworthy explanations.

\section{Distance Metric \& Clustering}
\label{sec:distancemetric}
\begin{figure}[tb]
    \centering
     \includegraphics[width=0.475\textwidth, trim={0.7cm 0.25cm 0.7cm 0.25cm}, clip,alt={The left tree splits on feature 1 first, and then on feature 2. The right tree splits on feature 2 first, and then on feature 1. All paths following this are hence incidentical.}]{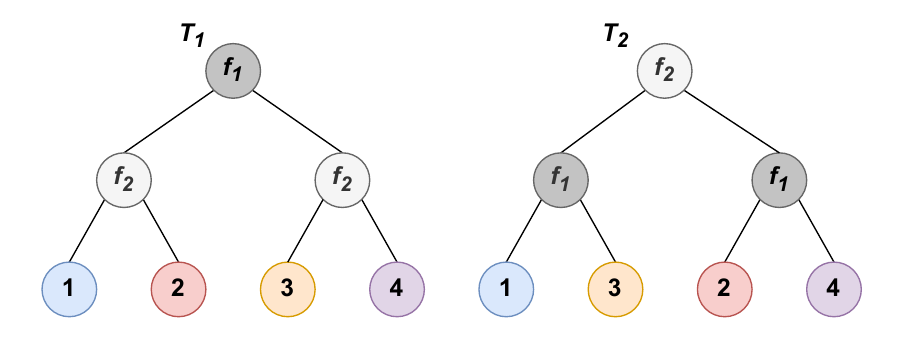}
    \caption{Two trees with a different order of split points but the same rules. Thresholds for features \(f_1\) and \(f_2\) are omitted for illustration purposes. Our distance metric would result in a distance of zero, while methods that only focus on the structure of the tree would give a non-zero distance\cite{laabs2023identification, eirich2022rfx}.
    }
    \label{fig:distance}
\end{figure}

Our idea is to use cluster-based visualizations that \revision{visually} aggregate information from several decision trees.
For this, we draw on the concept of representative trees in order to reduce the amount of information that is visualized at once and help to make sense of a large number of trees.
Before visualizing the random forest model, a suitable distance metric is needed~\cite{chipman1998extracting} to compute the clusters and select representative trees for each cluster. 

As a decision tree can be represented with a set of rules, we decided for a rule-based approach that incorporates the thresholds that are used in the decision trees, \revision{ignoring} their (path) order.
Figure~\ref{fig:distance} shows an example of why the order of the nodes (topology) is less important than the rules.
Our approach ensures that similarity reflects on the decision logic, and not only the used features and the used predictions.
Furthermore, we only compare rules that lead to the same predictions to include the semantic component and to ensure that the distances between trees correspond to meaningful differences in decision making.

The distance between two decision trees \(T_1\) and \(T_2\) is computed as the average rule distance:
\begin{equation}
d(T_1, T_2) = \frac{1}{|R(T_1)|} \sum_{r_i \in R(T_1)} \min_{r_j \in R(T_2)} d_R(r_i, r_j)
\end{equation}
where \( R(T_1) \) and \( R(T_2) \) are the set of of rules (paths) from trees \( T_1 \) and \( T_2 \), respectively \revision{and \(|R(T_1)|\) is the number of rules in \( T_1 \)}.  
For each rule \( r_i \) in \( T_1 \), we find the closest matching rule \( r_j \) in \( R(T_2) \) that predicts the same class, using the distances between the rules \( d_R(r_i, r_j) \). 
% Rule Distance Equation
For a given rule pair \( r_i \) and \( r_j \), the rule distance is the average interval distance across all features:
\begin{equation}
d_R(r_i, r_j) = \frac{1}{F} \sum_{f=1}^{F} d_f
\end{equation}
where \( F \) is the number of features. 
If there is no rule that predicts the same class, a maximum distance of 1 is assumed for \(d_R\).
Let the intervals for a feature \( f \) in two rules be denoted as:
\[
I_1 = [a_1, b_1], \quad I_2 = [a_2, b_2]
\]
The interval distance \cite{rico2022similarity, bouchet2023measures} for a feature \( f \) is defined as:
% Interval Distance Equation
\begin{equation}
d_f = 1 - \frac{|I_1 \cap I_2|}{\max(b_1 - a_1, b_2 - a_2)}
\end{equation}
The result is a vector for each tree that contains the distances to the other trees as its elements. 
\revision{We also tested methods based on prediction agreement, split points and structural similarity \cite{chipman1998extracting, banerjee2012identifying,laabs2023identification,bakirli2017dtreesim} but the dendrograms for our test data did not have strong cluster separation or were influenced by outliers. Furthermore, while plotting the rules in 2D using their distances to each other and MDS, a good separation could be observed between the clusters of rules.} % images are in Figures/distance-clustering. dendrograms of all methods with our clustering on glass and mds plot of the rules with color=prediction label

For clustering, \revision{we experimented with density-based methods like DBSCAN and HDBSCAN, but these approaches failed to detect meaningful clusters, likely because the vectors are derived from a distance matrix. Consequently, a distance-based clustering method seemed more fitting. Furthermore, with density-based methods, not every point is always assigned to a cluster, which is required for our approach.
}

We first calculate a complete linkage matrix as an agglomerative hierarchical clustering method\cite{clustering}. This ensures that all points within a cluster are relatively close to each other, leading to compact clusters. \revision{Additionally, we tried single linkage, but this led to the chaining effect, resulting in less satisfactory dendrogram cuts.}
The result \revision{of the linkage} is a dendrogram, which is then used by the dynamic hybrid cut algorithm to decide the final cluster assignment\cite{langfelder2008defining}.
The advantage of dynamic hybrid cut is the usage of dynamic branch cutting instead of a constant height cutoff value for the dendrogram. This takes advantage of local properties in the branches and creates clusters automatically that would otherwise be only identifiable by visualizing the dendrogram.
In practice, we computed the dynamic hybrid cut on the dendrogram with varying minimum cluster sizes to enable users to interactively change the number of visible clusters.
Finally, we compute the representative tree for each cluster by choosing the medoid, which is the tree with the minimal distance to all other trees of the cluster. We use this representative tree as a basis to map the rules to in the Rule Plot (Section~\ref{sec:ruleplot}) and the first tree in Section~\ref{sec:treeview}.

\section{Visualization}
\label{sec:vis}
An overview of the system is shown in Figure~\ref{fig:teaser}. We explain the components of our system in order of decreasing aggregation. We start by explaining the components of the Sidebar with the global view of the dataset and model (Section~\ref{sec:sidebar}) as shown in Figure~\ref{fig:teaser}a. We then proceed with explaining the more detailed view of the clusters through the Feature Plot (Section~\ref{sec:featurePlot}) and Rule Plot (Section~\ref{sec:ruleplot}) shown in Figure~\ref{fig:teaser}b. Afterwards, we present a fully detailed view of the individual trees in the clusters (Section~\ref{sec:treeview}) shown in Figure~\ref{fig:teaser}c. We conclude by presenting the linked interactions present in the system (Section~\ref{sec:interactions}).

\subsection{Sidebar}
\label{sec:sidebar}
\begin{figure}[tb]
    \centering
     \includegraphics[width=0.49\linewidth, trim=0 31cm 0 8cm, clip]{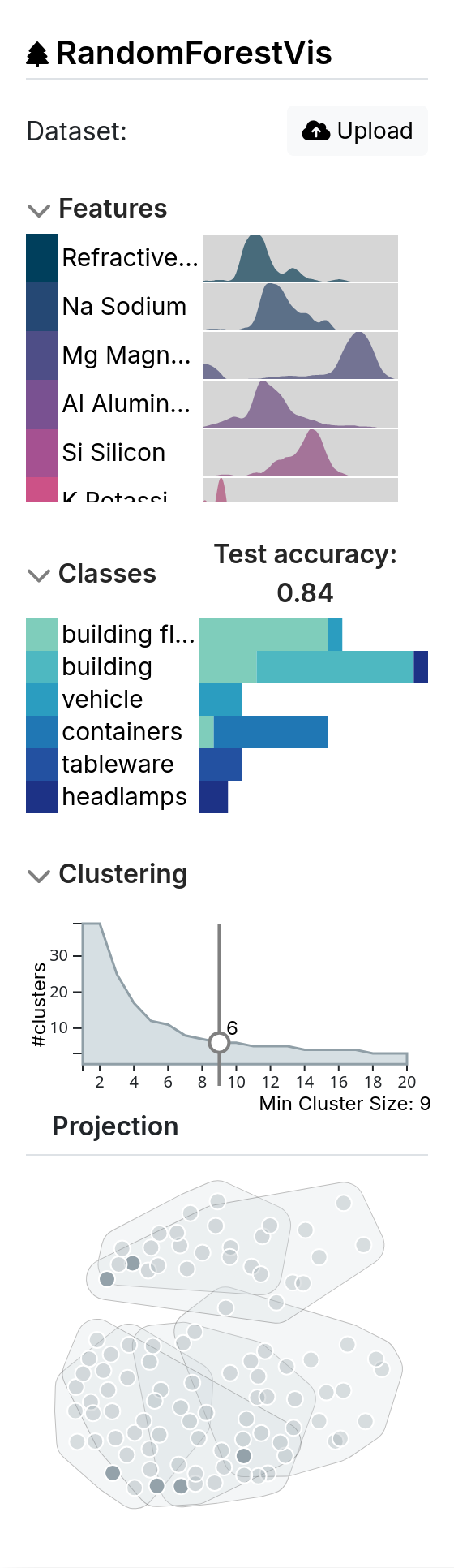}
     \includegraphics[width=0.49\linewidth, trim=0 2cm 0 36cm, clip]{Figures/Sidebar.png}
    \caption{The Sidebar of our system showing the ``Glass'' dataset, with scented area chart widgets for features (top left), classes and their (mis)classifications (bottom left), interactive cluster control (top right), and projection of the decision trees with representative trees for each cluster highlighted (bottom right).
    }
    \label{fig:sidebar}
\end{figure}

In the Sidebar, several visualizations are presented that provide an overview of the distribution of the underlying dataset and the overall model performance, as well as the clustering and distances between decision trees.

To visualize the distribution of features (Figure~\ref{fig:sidebar} top left), we use small area charts as scented widgets~\cite{willett2007scented} with kernel density estimation using the Epanechnikov kernel~\cite{epanechnikov1969non} to estimate the feature distributions. Users can filter the value range to explain the behavior on specific input clusters of interest (\textbf{T4}).
For each plot, a bandwidth is selected based on Silverman's method~\cite{silverman2018density} to prevent charts that are too noisy or too smooth. 
For categorical features, a histogram shows the value distribution. As our design requires both colors for the features and the classes, and there can be many features and classes present, we cannot easily use a categorical color scheme \revision{because of limitations in the number of distinguishable hues and reusing colors between features and classes would risk confusion and reduce clarity}. Hence, we instead use two sequential color scales with different hues for the features and classes, and use the same ordering for the features here as in the Feature Plot (see Section\ref{sec:featurePlot}) to reduce the difficulty of identifying features.

To visualize the model performance in general (\textbf{T3}), we use a classification matrix (Figure~\ref{fig:sidebar}, bottom left) and display the overall accuracy. Similarly to the features, using a categorical color scheme is not feasible here, and we use a sequential color scale of blue hues to minimize confusion with the colorscale of the features.

To enable users to change clustering, we use a scented area chart widget to display how the number of clusters depends on the minimum cluster size parameter of the clustering algorithm. Users can select the desired aggregation level with direct manipulation using a slider. By default, this is set using an approach similar to the elbow method by detecting where the maximum change in slope is and then selecting the subsequent point where the slope change reaches minimum. 
Linked to this slider is the projection view that shows a two-dimensional Multi-Dimensional Scaling (MDS) plot of the decision trees. 
We chose MDS because it allows us to visualize the trees based on their pairwise distances. This also makes the data easier to interpret, as the plot \revision{truthfully} reflects the relative distances between the trees.
For each cluster, the convex hull is computed, and the representative tree is highlighted.

Finally, users can select either the (``Glass''~\cite{glass_identification_42} or ``Penguin''\cite{penguin}) dataset that will be used in the user study, or upload their own dataset. When uploading a dataset, it is automatically processed using the standard random forest algorithm of scikit-learn~\cite{scikit-learn} and visualized. 

\subsection{Feature Plot}
\label{sec:featurePlot}
\begin{figure}[!tb]
    \centering
     \includegraphics[width=\linewidth,alt={Each level is visualized as a row, with the width of the row mapping to the percentage of times a certain feature is used on this level. The features are colored to distinguish between them.}]{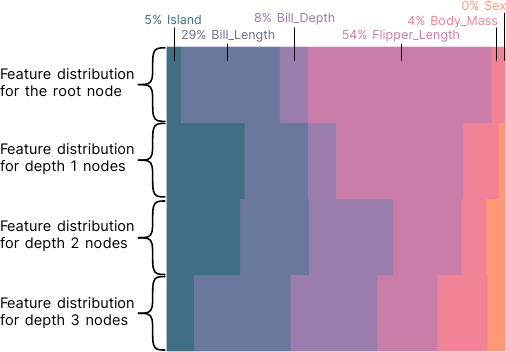}
     \caption{The Feature Plot visualizes the distribution of features in a (cluster of) tree(s) per level. The most important features are automatically placed near the top as these are split on first by decision trees.}
    \label{fig:featureplot}
    \vspace{-1em}
\end{figure}

After grouping the decision trees into clusters, users need to be enabled to analyze the clusters in a concise and comprehensible manner. For the visualization of the cluster of trees, we need a scalable visualization that both supports individual trees, and a cluster of multiple trees. In general, to understand the behavior of the random forest, we need to explore and understand the high-level characteristics of the clusters. To this end, the tree-structure of the individual trees is less important, but we rather want to understand for each cluster:

\begin{itemize}[noitemsep]
\item what features are used in the trees (\textbf{T1}, \textbf{T4});
\item what is the importance of each feature (\textbf{T5});
\item how similar are the features used (\textbf{T3}, \textbf{T4}); and
\item how complex/deep are the underlying trees (\textbf{T1});
\end{itemize}

\noindent For this, we designed a novel visualization: the Feature Plot (see Figure~\ref{fig:featureplot}).
Each row in a Feature Plot represents one level of the tree(s). The number of rows is therefore equal to the maximum depth of the contained trees. This provides a quick overview whether the trees in the cluster are shallow (few rows) or deep, hinting at the complexity of the trees. For each level, we determine the frequency of the features that are used in the underlying trees. We then divide the rectangular row into smaller rectangles, one for each unique feature used, and scale these horizontally proportional to their frequency. Each rectangle is colored according to the associated global feature color. This process is repeated for each level. For each level in the Feature Plot, the order of the features is consistent to enable comparison and quick lookup. 

This Feature Plot aggregate visualization \revision{provides additional cluster information compared to feature importances alone: 1) it
shows what features are used in the tree most (size of the rectangles), 2) what their importance is (higher features are considered more important, as they split the dataset early on and thus have a higher discriminative power), 3) how similar the features used are (uniform colors or different ones throughout the levels), 4) how complex they are (many different features at the levels or just a few), and 5) how the contained trees are shaped (equal depths or varied).} Many interesting insights can be gained from the plots; we expect a limited number of features at the higher levels, agreed upon by the underlying trees. More features in the middle levels, as diversity grows with each level. And few features at the lowest levels suggest that there are only a few specialized (overfitted) deeper trees. Additionally, we can derive what features are important at each level, and quickly identify whether features are used at all.
We layout the Feature Plots in a (rectangular) grid and sort them based on cluster size, starting with the largest cluster. This provides a quick overview of both the number of clusters as well as their size. 
%Hovering over a feature allows for \interaction{highlighting} of this feature.

\subsection{Rule Plot}
\label{sec:ruleplot}

\begin{figure}
    \centering
    \includegraphics[width=1\linewidth,alt={Each row in the feature plot show a single feature. A rule is a single vertical column. Within cell, rules are overlaid using opacity to create areas of darker and lighter shades. Dark shades indicate most rules include this part of the range. White areas include no rules use this part. In the example given in the figure, bill length is always above 30 percent for Gentoo Pinguins. At the very bottom of the chart, the classification matrix is displayed. Here, we see a few misclassifications for representative rules, with Gentoo being classified as Adelie.}]{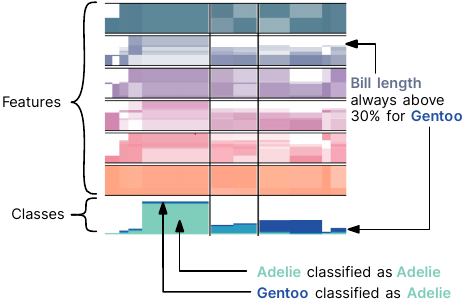}
    \caption{A Rule Plot that shows the decision trees in one cluster. Aggregated decision rules are displayed vertically along their features, with the classification matrix for the rule at the bottom.}
    \label{fig:ruleplot}
\end{figure}

To visualize the decision rules present in a cluster of trees, we use a second novel visualization: the Rule Plot (see Figure~\ref{fig:ruleplot}). This allows us to visualize the overall decision process in the model (\textbf{T1}), as well as to understand the decision process for a particular rule (\textbf{T2} after \interaction{filtering}).
As the number of decision rules in a decision tree is equal to the number of leaves, even for shallow trees, it is infeasible to visualize all decision rules \revision{individually. Instead, we map the set of rules $R(c)$ within a cluster onto a representative tree $T_r$ that is the medoid in the cluster, and visualize all rules through this representative. This allows us to show all rules in the decision trees in this cluster using visual aggregation, in contrast to simplifying to a single decision tree or showing only a representative, which would inherently lose some rules.}

Let $R(T)$ be the set of decision rules for tree $T$. Then, for each tree $T$ in a cluster \revision{$c$}, we map each decision rule $r \in R(T)$, towards the closest decision rule $r^* \in R(T^*)$ of the representative tree $T^*$ of the cluster \revision{$c$}.
The distance from rule $r$ to $r^*$ is defined as in Section~\ref{sec:distancemetric}. As an end result of this process, every rule $r$ in our cluster \revision{$c$} is mapped to one of the rules \revision{$r^*$ of the representative tree $T^*$}, with $R(r^*)$ the set of decision rules mapped to $r^*$. 

Using an appropriate distance metric is crucial as otherwise the visualization of these aggregations will not yield useful results. The most important factor is that we are \revision{mapping} rules resulting in the same output classification, as otherwise we need a much more complicated visualization design for the \revision{mapping} to split the differences apart. Secondly, we would like the rules to be similar in nature, i.e., have similar ranges for each feature values. This relates back to our choice for the distance metric used for distances between rules in Section~\ref{sec:distancemetric}. 

Using the rule mapping, we then visualize the aggregated rules $R(r^*)$ for each rule $r^* \in R(T^*)$ in the representative tree $T^*$ using a vertical column. We place the features vertically in the same order as in the Sidebar. For each feature, we \revision{use} a vertical heatmap \revision{to visualize} how often each part of the feature range is used in $R(r^*)$ using opacity blending. \revision{The darker the area, the more rules in $R(r^*)$ use this part of the feature range. In \autoref{fig:ruleplot} we see an example of this in the \billlength{bill length} where all involved rules for \gentooColor{Gentoo} are above the 30\% mark, and even darker areas above the 70\%.}
Below the vertical heatmaps for the features, we use a stacked-bar chart of the classification results of $R(r^*)$ for the test data to complete the Rule Plot. 

We use the width of the heatmaps to highlight those rules that are most often applied by scaling them based on the number of classifications made with this rule. We ensure that lesser used rules remain visible by capping the scaling at a factor 10. We place the rulesets $r^* \in R(T^*)$ side by side, sorting the ruleset using a 1-dimensional embedding of the distances between the rules of the representative trees $T^*$. This ensures that similar rules are close to each other, allowing for easier pattern identifications. Black vertical bars are positioned between groups of rules with different classification outcomes, to visually group these together and make it easier to identify the groups.

\subsection{Decision Trees within Clusters}
\label{sec:treeview}
\begin{figure}[tb]
    \centering
     \includegraphics[alt={A decision tree is visualized as a modified sankey diagram. A smoothed distribution chart is used to visualize the range of values for a feature in a node, with a black vertical line to indicate the split point. Values that are no longer relevant to the previous split points are grayed out in the distribution chart. The sankey lines split further from this line, indicating which part of the decision trees goes further to the left and to the right. At the leaf nodes, the classification matrix is shown using a simple horizontal stacked bar chart, using colors to distinguish between the classes.}]{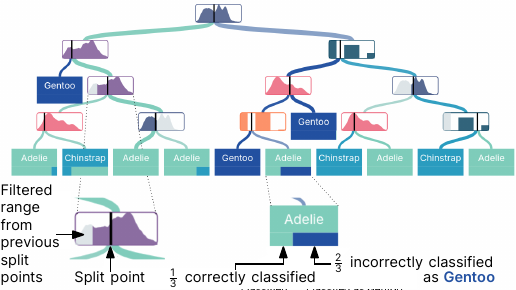}
    \caption{A single decision tree from the ``Penguin'' dataset. Internal nodes show the active feature distribution and the split points. Leaf nodes show the classification matrix for the decision path.}
    \label{fig:node-link}
\end{figure}
To provide a detailed view of the individual decision trees, we visualize all decision trees as node-link diagrams starting with the representative tree for a selected cluster (See Figure~\ref{fig:teaser}c).
For this, we use the enhanced Reingold-Tilford algorithm for variable-sized nodes\cite{van2014drawing}.
As decision trees are often visualized as node-link diagrams, this view should help users become familiar with the visualization system \revision{and help} the learning process.
For each tree, the accuracy of the training and test data is visible together with a tree number.
We use both training and test data to prevent unused paths in the decision trees.

An example decision tree explaining the node design \revision{is shown} in Figure~\ref{fig:node-link}.
The nodes \revision{are either} feature split points and class predictions. 
For each split point, we visualize the node as a kernel density estimation (KDE) plot of that feature together with a black line showing the split point.
If a feature appears multiple times along a path in the decision tree, the previous split points are used to color the intervals outside the current range in gray.
The prediction nodes show the target class's text label together with a stacked bar that shows the (miss)-classification through class colors.
We use a KDE to visualize the split value to be in line with the presentation of the features in the feature chart, to be consistent with the colors in general, and to give an explanation how the decisions made relate to the distribution of the features.

The edges are drawn using Bézier curves to support easier line following through continuity. \revision{An edge} starts at the split point of the parent and ends at the center of the child.
Each edge has a thickness \revision{corresponding} to the number of samples in the dataset that follow this path in the tree (similar to~\cite{Elzen2011BaobabView}). 
The color of an edge is based on the classes of samples that pass through \revision{it}. This enables users to quickly see which paths are important and what is the dominant class of a path.

\subsection{Interactions}
\label{sec:interactions}
As previously presented, our system consists of multiple linked views showing the overview of the data in the Sidebar, a more detailed clustered view in the middle, and the decision trees themselves on the right. We have linked these views using filtering and brushing (highlighting). A video of all interactions is available in the supplementary material.

\begin{figure}[!t]
    \centering
    \includegraphics[width=1\linewidth,alt={A unfiltered view of a cluster (Feature Plot and Rule Plot) is shown on the left, the filtering interface in the middle which works via brushing, and the filtered view of the cluster on the right.}]{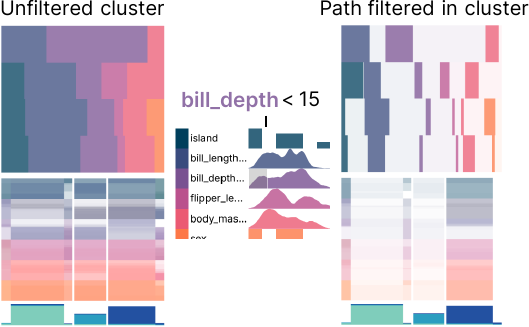}
    \caption{Filtering based on \billdepth{bill depth} below 15. In both the Feature Plot (top) and Rule Plot (bottom), decision rules and path that cannot have \billdepth{bill depth} below 15 are hidden. In the Feature Plot we see the updated distribution of features,  showing that all features remain relevant for the classification. In the Rule Plot we see the majority of penguins with small \billdepth{bill depths} are from the \gentooColor{Gentoo} species, although there are a few from the \adelieColor{Adelie} species as well.}
    \label{fig:filteringExample}
\end{figure}

\noindent\textsf{\textbf{Filtering}}
Through filtering the dataset, users can zoom in on particular feature values and (mis)classifications of interest to better understand (\textbf{T1,T2,T4, T5}) the decisions made by the model.
For every decision rule, we can filter on three different components: (1) The value ranges for each feature that are valid within this rule, (2) The output classification of this rule, and (3) Misclassifications that this rule has made. 
For (1), we filter the data using the Distribution plots for the features in the Sidebar. Users can drag the areas to select the value ranges of interest for each feature. We then filter out a decision rule $d$ if there can be no instance that satisfies the selected value ranges.  
For (2) and (3) we combine the selection of it using the classification matrix in the Sidebar. Users can select specific (mis)classifications or an entire class to select all classifications by clicking on them. We then filter out a decision rule $d$ if the selected (mis)classifications are not present in $d$. 

Once we know which decision rules should be visualized, we can update the other visualizations to highlight this. 
For the decision tree visualization, this is simple as we simply fade out any paths corresponding to inactive decision rules, corresponding to standard dimming, to retain the overview.

For the Rule Plot, we do not visualize any of the decision rules that are filtered out. Fading here is not feasible, since opacity is used as a visual variable. After filtering for the Rule Plot, there is a choice to re-normalize the data in the vertical heatmap. The advantage of this would be that the filtered rules would have maximal visibility. The downside is that filtering a value range would paradoxically increase the visibility of the least-specific rules, as these are the last ones to be filtered out. Hence, we decided to not re-normalize the data, and keep the filtering more intuitive and stable at the cost of slightly lesser visibility. An example of this filtering is shown in Figure~\ref{fig:filteringExample}.

In the Feature Plot, we are not visualizing the rules directly. Instead, we visualize how often features are used per level $l$ of the tree, which information is not contained in a decision rule as it is topology unaware. Hence, we filter based on the decision paths in a cluster. 
Similarly to decision rules, we can filter out decision paths that do not lead to the selected (mis)classifications. 
For the feature values, for each node $v$ at depth $l$ in a decision path $p$ we consider the decision rule $d$ formed from the root of $p$ to $v$, and filter in or out this node based on $d$. 
As decision rules only get stricter the further down $p$ we go, this ensures a consistent filtering strategy of the nodes by showing subpaths starting from the root. For each level $l$ and feature $f$, we calculate the percentage of nodes in the decision paths at depth $l$ that are filtered. We then visualize this percentage using a white area on the feature to be consistent with the Rule Plot filtering within this interaction. An example of this filtering is shown in Figure~\ref{fig:filteringExample}.

\noindent\textsf{\textbf{Highlighting}}
Hovering over a feature in any of the views highlights this feature in all views. Aside from the advantage of being able to see the feature of interest clearly in all views, clusters, and decision trees in one go, it also helps alleviate some of the issues with choosing a quantitative color scheme for categorical data. By highlighting the feature everywhere, users can quickly see which feature is which.

In most plots, highlighting is achieved by reducing the opacity of all other features. In the Rule Plot, this, however, is not an option, as opacity is already used in the vertical heatmaps as a visual variable. Instead, we increase the width of the black horizontal bar above and below the feature to highlight it (see the black bars around \calciumColor{Calcium} in the figure on the right).

\noindent\textsf{\textbf{Other interactions}}
Aside from the linked interaction, several other interactions in the system are available. We briefly list these here.

\textbf{Select level of aggregation} In the Sidebar users can select the required level of aggregation. After selecting the desired aggregation level, new clusters are generated and visualized in the middle view using the Feature and Rule Plots. Similarly, the decision trees of the clusters are updated as well.

\textbf{Cluster selection} By selecting a Feature or Rule Plot within a cluster, users are shown the decision trees for this particular cluster. 

\textbf{Projection hovering} By hovering over a point in the projection view in the Sidebar, all points corresponding to decisions trees in the same cluster are highlighted.

\textbf{Zooming and panning} In the Rule Plot and Feature Plot view, as well as the decision tree view, zooming and panning are implemented to allow for details to be more easily perceived.

\textbf{Tooltips} Tooltip information is available for the classification matrices showing exact values (including those in the Feature Plot), for the distribution plots showing exact values, split points on the decision trees, and the number of samples for an edge in a decision tree. These enable users to obtain the exact information where relevant.

\section{Evaluation}
\label{sec:evaluation}

To evaluate our approach, we follow standard approaches for evaluating explainers for random forests \cite{chatzimparmpas2023visruler,zhao2018iforest,eirich2022rfx}, by presenting a case study to demonstrate how our approach resolves the tasks in Section~\ref{sec:tasks}. In addition, we conducted a small qualitative user study to determine the usability of the system and identify areas of improvement.

%[1] Re-intepreting rules. Cardiocartograph, HELOC and Nomad 2018 datasets. 
%[3]: ? No longer have access
%[4]: Iris and breast cancer wisconsin datasets. Code available.
%[14]: Visruler. Proprietary data. Code available
%[17] RFX: Proprietary data. No code available.
%[34] Random Forest Similarity Maps. Iris, breast cancer wisconsin, US Election datasets , no code available.
%[35]: Understanding the forest: Proprietary data. No code available.
%[38]: Rulematrix: breast cancer wisconsin  and Pima Indian Diabetes datasets. code available.
%[39]: Treepod: UCI Census Income dataset. No code available.
%[56]: Baobabview: Image segmentation and breast cancer Ljubljana datasets. 3rd party code available.
%[61]: Timbertrek: Compas dataset. Code available.
%[62]: Forest Floor: White wine quality and Contraceptive method choice datasets. Code available.
%[65]: IForest: Titanic dataset. No code available.

%There is no standard baseline dataset that is used for comparison purposes. Similarly, the tasks evaluated in the literature differ depending on their intended use, making a direct comparison difficult. 

\revision{Most papers explaining random forest use different datasets, and hence there is no standard baseline dataset to compare with.} For our evaluation, we picked two different datasets: (1) The ``Penguin''~\cite{penguin} dataset cut off at a depth of 4 to intentionally introduce misclassifications. \revision{We use this} in our user study as a training dataset due to its limited complexity (3 classes, 6 features, clear distinction between classes with 99\% accuracy). (2) The ``Glass''~\cite{glass_identification_42} dataset, which is a more complicated dataset with 6 classes, 9 features, and significant overlap between classes (84\% accuracy from the random forest). The complexity of the dataset should give us a representative indication of the efficacy of our approach on real-world datasets. \revision{Additionally, it is a well-known, open-source and standard dataset that sees active use, which ensure ease of reproducibility and comparability. Finally, both the number of features and classes, while larger, are still within the scalability capabilities of our approach.}

\subsection{Case Study}
To demonstrate how our approach can resolve the tasks in Section~\ref{sec:tasks} we perform a case study using the \revision{well-known} ``Glass'' \cite{glass_identification_42} dataset\revision{, highlighting what the user can do with the insights}. A complete view of our system is shown in Figure~\ref{fig:teaser}.

\begin{wrapfigure}[9]{r}{0.5\columnwidth}
\vspace{-1.7em}
  \centering
    \includegraphics[width=\linewidth,alt={A snapshot of the sidebar which shows the classification matrix. No caption.}]{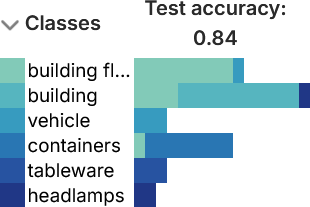}
\end{wrapfigure}
We start by evaluating how well the model works globally using the classification plot (\textbf{T3}). Most of the classes are classified well (84\% accuracy), but for class \buildingColor{Building} a significant fraction (25\% as shown on hover) of the instances are misclassified into class \buildingFloatColor{Building float} (\textbf{T4}). \revision{\textbf{Insight:} Combined with the imbalance in class frequency, this shows the modeler which additional data is desired in case the model is unsuitable or needs to be improved for their purposes.}

\begin{wrapfigure}[15]{l}{0.4\columnwidth}
\vspace{-1.3em}
    \centering
    \includegraphics[width=1\linewidth,alt={A snapshot of one cluster in the ruleplot showign Building and Building float. No caption.}]{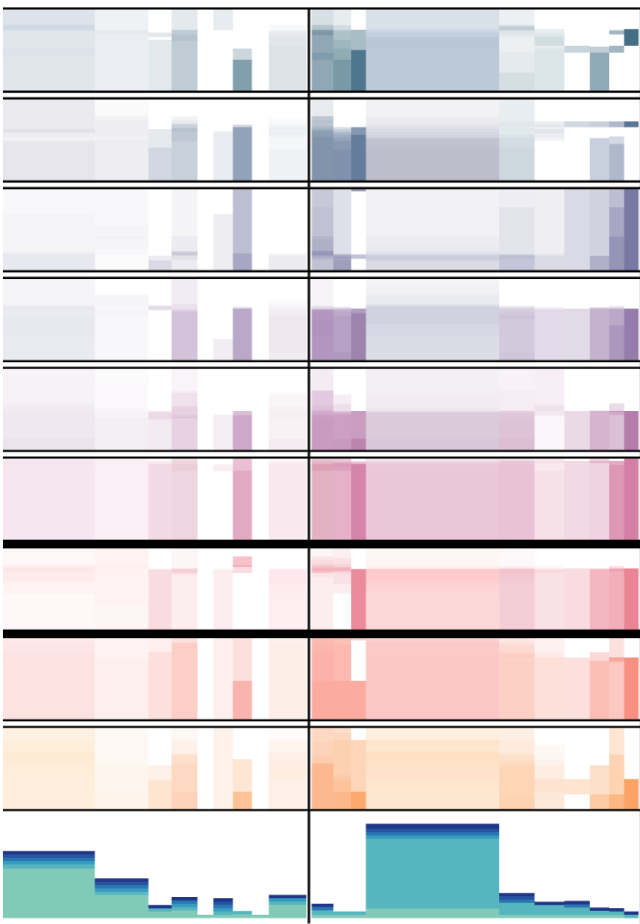}
\end{wrapfigure}
We then investigate deeper into how the classification \buildingColor{Building} is determined and their misclassifications (\textbf{T1, T5}). We filter the classification of \buildingFloatColor{Building float} to its misclassification \buildingColor{Building} and the correct classification \buildingFloatColor{Building float}, which results in two visible columns per cluster in the rule plot. \revision{One such cluster is shown on the left.} In the right grouped column, we see the general classification rules used to correctly classify \buildingColor{Building}, rules that do not follow this classification are faded out. For some features, there are strong split criteria with a vast majority of the rules and classifications made using a similar threshold (i.e. the highlighted featured \calciumColor{Calcium} for values of 8.3).
In the left grouped column, we have the misclassifications from \buildingFloatColor{Building float} to \buildingColor{Building}. Rules that do not have this misclassification are faded out. From the classification barcharts at the bottom, we see that this misclassification is made in most rules, although not by significant amounts in many of them. We also see that many of the rules where misclassifications are made follow quite similar patterns to those of the correct classifications. This indicates that there are only small differences between these two classes, and that it is a combination between multiple features. \revision{\textbf{Insight}: With additional domain knowledge, users can determine if there is indeed a blurry line between the two classes, or whether additional features could be introduced to make this distinction more clear. Moreover, it tells them that classifications between these two classes are quite sensitive to noise.}

\begin{wrapfigure}[10]{l}{0.4\columnwidth}
\vspace{0em}
    \centering
    \includegraphics[width=1\linewidth,alt={A snapshot of the Featureplot for one cluster after filtering. No caption.}]{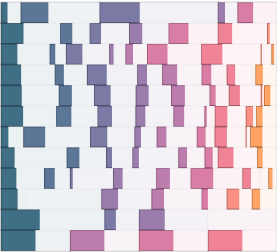}
\end{wrapfigure}
\begin{figure}[h]
    \centering
    \includegraphics[width=1\linewidth,alt={A snapshot of three different clusters from the Featureplot+Ruleplot. Using the presence or non-presence of features are the higher levels, features of importance can be identified.}]{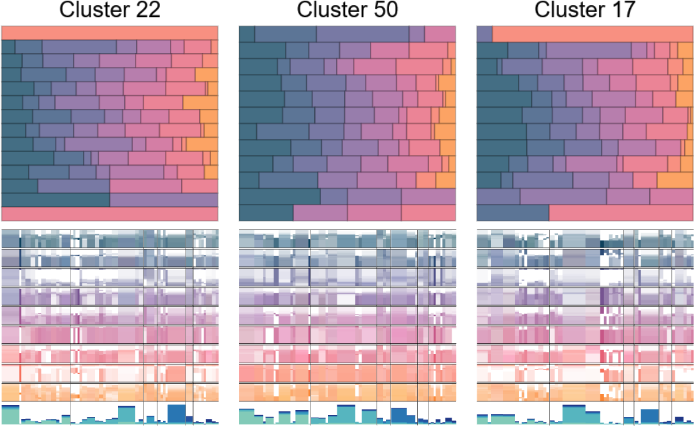}
    \caption{The clusters with different features of importance shown.}
    \label{fig:unfiltered}
\end{figure}

Continuing by looking at the filtered Feature Plots of \revision{cluster 50} for these two classifications, we can see which features are being split and how high in the tree. We notice that all features are present, but some only become relevant much deeper \revision{(i.e. \refractiveindexcolor{Refractive Index}, first feature, and \ironColor{Iron}, last feature)} indicating that this is a finer split to distinguish between two final classifications. \revision{\textbf{Insight:} In contrast to the two clusters in~\autoref{fig:unfiltered}, there is more variety in splits in cluster 50. This indicates that many trees are substantially different, and there is difficulty in generalizing a classification for these specific two classes. This could also indicate that one cannot easily remove any feature from the model as all features are used. Hence, using less features to reduce the cost of data collection could come at a cost in distinguishability between these two classes.} 

We go back to the unfiltered view (see Figure~\ref{fig:unfiltered}) of the data to determine the general features that are most important for the model.
In the Feature Plot, we see that three different features are considered important. \bariumcolor{Barium} is always used as the first split point in two of the clusters, while in the other cluster it is much more balanced with \magnesiumcolor{Magnesium} and \sodiumcolor{Sodium} also being often present in all trees. We can also see this split in the \revision{rule plot of the cluster without \bariumcolor{Barium} as the first split point.} The rules are much more complicated than in the others, as barium is used to split much later.

\begin{wrapfigure}[7]{r}{0.5\columnwidth}
\vspace{-1em}
    \centering
    \includegraphics[width=1\linewidth,alt={Part of a single decision tree. It is highly unbalanced, with one of the two branches immediatly going towards a leaf node for most internal nodes.}]{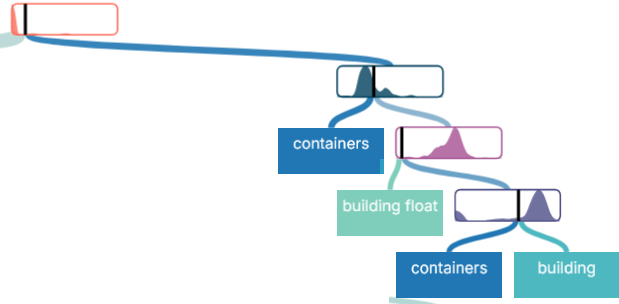}
\end{wrapfigure}
\revision{We can validate this by looking at the node-link view, where for the clusters with \bariumcolor{Barium} as a split point, many trees make classifications at a low depth with perfect accuracy. In contrast, for the cluster where \bariumcolor{Barium} is not present in the first level of the Feature plot, these immediate classifications at low depth are not made. \textbf{Insight:} This gives the user insight in the different ways the classifications are being made by the random forest and shows the simple and complex cases. } \\

\begin{wrapfigure}[12]{l}{0.16\columnwidth}
\vspace{-1.2em}
    \centering
    \includegraphics[width=1\linewidth,alt={A Rule Plot filtered to only show a very narrow range of feature values. This results in a series of vertical bars that show which rule is usually active.}]{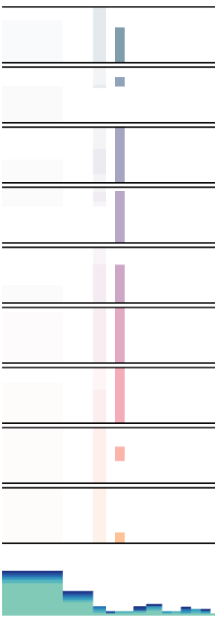}
\end{wrapfigure}
We continue by drilling down on a specific instance, wanting to know how this obtained a particular classification \textbf{(T2)}. We input the feature values of an instance in the filtering, and see the rules being used for this instance. In addition, we can see how sensitive these rules are to changes, i.e. how much of a permutation is required before a different rule would need to be followed. 
In general, for a single instance, multiple rules can be active due to the nature of the random forest. While the Rule Plot visualization makes it possible to view the active rules for an instance, it is still somewhat difficult to view several rules in different clusters and compare them. \revision{\textbf{Insight:} For this particular case, we see that most trees capture it using a very similar rule that has narrow ranges for \sodiumcolor{Sodium} (2\textsuperscript{nd} row), \bariumcolor{Barium} (8\textsuperscript{th} row) and \ironColor{Iron} (9\textsuperscript{th} row). This indicates that the classification of this instance is quite sensitive to changes in these values.}

From the case study, it can be seen that most tasks can be performed with the system, and they lead to non-trivial insights into the data. The various visualizations allow for both detailed views as well as keeping the general overview, and the linking between them through interaction is crucial to control the desired aggregation level for the task.

\subsection{User Study}

In addition to the case study, we perform a small user study to gain an indication about the usability of the tool using a think-aloud methodology. For this we recruited two domain experts, who had not been involved with the tool in any way. Both participants regularly work with machine learning tools, with one participant actively working with random forests.

Before the study started, the participants were informed about the goal of the study and were asked to fill in a consent form for permission to use their responses in this study. It was made clear that consent could be withdrawn at any point during the study.

We then start the study with a short reintroduction of random forests and explain how the various components of the system work, before asking the participants to explore the ``Penguin'' training dataset.

After this initial exploration, we switch to the ``Glass'' dataset, and ask the participants to complete the following tasks sequentially.
\begin{enumerate}
\item What are the most important features in the model? \textbf{(T5)}
\item How well is the model performing? \textbf{(T3)}
\item Which class is misclassified the most? \textbf{(T4)}
\item How does the model decide that an instance belongs to class \headlampsColor{Headlamps} \textbf{(T1+T4)}
\item What are the factors resulting in misclassification of \buildingColor{Building} \textbf{(T4)}
\item How representative are the clusters shown of the underlying random forest?
\end{enumerate}

After the participant finished all tasks, we ask them for open-ended feedback on the tool. Furthermore, we ask them to complete the SUS survey\cite{brooke1996sus} and ICE-T~\cite{wall2018heuristic} survey to gain additional information on the user-friendliness and usability of the tool as a whole. The anonymized notes taken, as well as the SUS and ICE-T scores, are available as supplementary material.

\subsection*{Results}

The domain experts were able to make sense of the interactions within the system, and these helped when trying to understand the relationship between all the views. Especially the linking from the features to the Feature Plot, Rule Plot and decisions trees was found helpful, as this helped clarify how the components were linked and gave a clear reference for the colors of the features and classes.

\smallskip
Both participants indicated that the variety of different features and views were useful to them. They also mentioned that there were a lot of different views to keep in mind and remember, and felt they still have things to learn from the system by the end of the session. 

\smallskip
The decision tree view was found useful for both participants, as it was a more familiar view and easier to comprehend. Both participants used it extensively to validate their findings and generate potential hypotheses. Showing the classification matrix in the leaf nodes and the feature distributions in the internal nodes did not appear to make it more complex to comprehend, and was found useful in specific cases. 

\smallskip
The initial three tasks were all able to be resolved well by the participants. All views within the system were used for these tasks. For determining (mis)classification (4,5), the participants found multiple different factors, but since there is no clear-cut answer involving only one feature, they were unsure whether they had sufficient information.

For the last task (6), the more novice participant was unsure how to assess how representative the clusters shown were of the underlying random forest, as they had not fully internalized all the relations between the views. The other participant did not have such issues and was able to use all views well to answer this question. Both participants initially struggled to understand how the clustering was done in the Sidebar view. While they managed to change the aggregation level and discover how it worked, it was not intuitive to them, indicating that interaction for the level of aggregation selection could be improved. 

\smallskip
The participants also struggled with the size of the classification matrix underneath the clusters. While it was successful in indicating the major classes for a ruleset, obtaining more details (in particular exact numbers) by hovering over the rule was difficult due to the small size of the matrix without zooming in excessively.

\smallskip
Finally, the SUS survey results in a score of (70 and 82.5), placing it above the average of 68~\cite{sauro2016quantifying}. The ICE-T\cite{wall2018heuristic} which is more visualization specific gives a score of (6.02 and 5.65) out of 7. Together, these two surveys indicate that participants found the system usable and useful.

\section{Discussion and Limitations}
\label{sec:discussion}

In our user study, we noticed that the clustering interaction for the level of aggregation selection was not sufficiently intuitive. Our approach uses a hierarchical clustering method, which is well suited for an interactive system, but the method of setting the threshold can be improved. For example, the possibility of manually setting the cuts in a dendrogram would give the user more control~\cite{vogogias2016mlcut} and insight into the method. However, this poses challenges for visualizing such a dendrogram in a scalable manner.

While the classification bar charts at the bottom of the Rule Plots give a good idea of the distribution of misclassifications, we also noticed that it is difficult to determine exact values. Alternative visualizations or interactions should be considered here to mitigate this effect.

Our current visualization approach allows the analysis of the model, but misses an instance-based perspective, such as tracing how the model behaves for a specific sample in the dataset. This could be resolved by highlighting a selected data sample in the visualizations. \revision{A cluster-based approach to visualizing random forests offers a middle ground between the extremes of local analysis and full aggregation. Compared to analyzing individual decision trees, clustering reduces cognitive load by grouping structurally and functionally similar trees, allowing users to identify representative patterns without becoming overwhelmed by the forest's complexity. This enables general insights into model behavior while preserving important variability across subgroups of trees. In contrast, aggregating all trees into a single summary tree offers simplicity and ease of interpretation, making it appealing for non-expert users. However, this simplification can obscure important variability and rare but meaningful decision paths. Clustering avoids this trade-off by preserving structural diversity within each group while still offering a compressed, interpretable view of the model. However, the cluster-based approach also introduces challenges. It depends on the quality of the clustering algorithm and the chosen distance metric, which may not always reflect meaningful semantic differences for all tasks. Additionally, users may still need to interpret multiple cluster representatives, which can introduce complexity, particularly if the forest exhibits high heterogeneity.
}

In terms of scalability, our approach scales well with regard to the number of trees. However, it is limited in terms of the number of features and classes as we are using color to represent these, \revision{which might hinder real-world adoption}. As the number of categories increases, distinguishing colors on a sequential scale becomes more difficult. Although highlighting and selection interactions help mitigate this effect, it remains a challenge for datasets with many features. A possible solution is feature aggregation or selection, which would result in some information loss. \revision{Further, custom colorscales could be used for a specific dataset.}
Additionally, performance issues may arise when trees become deep and complex. In this case, a progressive analytics approach~\cite{stolper2014progressive} could help by analyzing a subset of trees, providing a first insight into important features, rules, and the cluster.

Our system is specifically made for visualizing random forests, but the Feature Plot, the Rule Plot, and the decision tree visualization can also be used to visualize singular decision trees. The Feature and Rule Plot may be usable for other types of classification algorithms as well, as long as decisions paths (Feature Plot) or decision rules (Rule Plot) can be generated through the approach. The clustering approach itself can be extended to general ensemble techniques for classification as well, requiring only a distance metric and a way to visualize the aggregation.

\revision{In general, by visualizing the distributions, classification results, as well as the random forest, common fundamental mistakes such as using the wrong encoding for a feature (quantitative instead of categorical), using the wrong technique (Random forest not classifying it well enough, or a single tree being sufficient), or having the wrong features can be brought to light in an intuitive way. These are often simple issues to resolve when detected, but nonetheless have a major impact if they are not spotted.}

The field of random forest visualizations is wide, and an in-depth comparison of the different approaches with standard tasks and datasets is needed, as well as an insight-based evaluation for real-world datasets. Currently, such a comparison is not possible as the systems have been tested on too diverse tasks and datasets.
The situation is similar for distance metrics for decision trees.
Some studies compare them~\cite{laabs2023identification}, but to the best of our knowledge there is no in-depth comparison of different semantic, structural, and rule-based approaches.
In addition, while it is possible to construct edge cases for different metrics, it is unclear how often these cases appear in random forests trained on real-world data.

Finally, while we have not included it into our system, it is nevertheless of interest for spatial data to allow for spatial data selection methods. Currently, we use scented widgets with brushing for the feature selection, but alternative approaches such as selection on a matrix, parallel coordinate plots, or directly on a map could allow for a more nuanced filtering.

\section{Conclusion}
\label{sec:conclusion}

In this paper, we introduced a novel visualization approach to enhance the interpretability of random forests by clustering similar decision trees based on both their \emph{semantics} and \emph{structure}. For this, we introduce a new distance metric that can be used to determine decision tree similarity. 

Our approach allows users to gain insight into the overall behavior of the model without being overwhelmed by individual trees or relying on oversimplified summaries of the entire classifier. To support this, we proposed two new scalable visualization techniques: the Feature Plot, which reveals the hierarchical structure of feature importance across clusters in the forest, and the Rule Plot, which presents the decision-making logic within clustered trees. 

We implemented these in a visual analytics prototype solution, supporting the user workflow through linking and brushing of the different components. Furthermore, we show how the linked visualizations relate to common interpretability tasks. Our case and user study demonstrate the effectiveness of our approach in improving model transparency. Future work could explore refining the distance metric and clustering method to further aid model interpretability. Additionally, it would be interesting to explore how our visualizations and general clustering-based approach can be adapted to other rule-based and tree-based models. Overall, we believe that, with our approach, we have added another useful tool to the user toolbox to increase model transparency.

%% if specified like this the section will be omitted in review mode
\acknowledgments{%
The authors wish to thank Sem Lommers for initial exploration of the topic and interesting discussions. Stef van den Elzen is partially supported by AI4Intelligence with file number KICH1.VE01.20.011, partly financed by the Dutch Research Council (NWO). %
Christofer Meinecke acknowledges the financial support by the Federal Ministry of Education and Research of Germany and by Sächsische Staatsministerium für Wissenschaft, Kultur und Tourismus in the programme Center of Excellence for AI-research ``Center for Scalable Data Analytics and Artificial Intelligence Dresden/Leipzig'', project identification number: SCADS24B. Tatiana von Landesberger acknowledges the financial support by BMBF Project Risk Principe and WarmWorld.

}

% bibtex
\bibliographystyle{abbrv-doi-hyperref}
\bibliography{bibliography}

% biblatex with biber
% \printbibliography                

\end{document}